# The Strength of Nesterov's Extrapolation in the Individual Convergence of Nonsmooth Optimization

Wei Tao, Zhisong Pan, Gaowei Wu, and Qing Tao

*Abstract*—The extrapolation strategy raised by Nesterov, which can accelerate the convergence rate of gradient descent methods by orders of magnitude when dealing with smooth convex objective, has led to tremendous success in training machine learning tasks. In this article, the convergence of individual iterates of projected subgradient (PSG) methods for nonsmooth convex optimization problems is theoretically studied based on Nesterov's extrapolation, which we name *individual convergence*. We prove that Nesterov's extrapolation has the strength to make the individual convergence of PSG optimal for nonsmooth problems. In light of this consideration, a direct modification of the subgradient evaluation suffices to achieve optimal individual convergence for strongly convex problems, which can be regarded as making an interesting step toward the open question about stochastic gradient descent (SGD) posed by Shamir. Furthermore, we give an extension of the derived algorithms to solve regularized learning tasks with nonsmooth losses in stochastic settings. Compared with other state-of-the-art nonsmooth methods, the derived algorithms can serve as an alternative to the basic SGD especially in coping with machine learning problems, where an individual output is needed to guarantee the regularization structure while keeping an optimal rate of convergence. Typically, our method is applicable as an efficient tool for solving large-scale $l_1$-regularized hinge-loss learning problems. Several comparison experiments demonstrate that our individual output not only achieves an optimal convergence rate but also guarantees better sparsity than the averaged solution.

## I. INTRODUCTION

IN THIS article, we aim at regularized optimization problems arising in machine learning, where the objective function is the sum of two nonsmooth convex terms: one is the loss from the learning task and the other is a regularizer such as $l_1$-norm for promoting sparsity. For better understanding, we first focus on the constrained optimization problem,

$$\min f(\mathbf{w}), \quad \text{s.t.} \quad \mathbf{w} \in \mathbf{Q} \qquad (1)$$

where $\mathbf{Q} \subseteq \mathbb{R}^N$ is a closed convex set and $f$ is a convex function on $\mathbf{Q}$.

A popular method to solve (1) is the projected subgradient (PSG) algorithm. When $f$ is smooth, it was proven that no first-order method can converge at a rate faster than $O((1/t^2))$ by Nemirovski and Yudin in the early 1980's [18], where $t$ is the number of iterations performed by the algorithm. However, PSG only has a suboptimal $O((1/t))$ rate of convergence [30]. This created a gap between the guaranteed convergence rate of PSG and what could potentially be achieved. Since then, various efforts have been paid to close this gap (for instance, see [30] and references therein). One well-known strategy is to perform the extrapolation operation, where momentum terms involving the previous iterations are added to the current iteration. Such technique dates back to the pioneering work of Nesterov in 1983 [19]. Specifically, Nesterov showed that the extrapolation step with suitably chosen parameters can accelerate the rate of convergence from $O((1/t))$ to $O((1/t^2))$, that is, this strategy makes PSG optimal among first-order techniques that can access only sequences of gradients [20]. So far, there have been many papers devoted to extending Nesterov's accelerated method to stochastic settings for regularized learning problems, where the regularization structure (e.g., sparsity, low-rank, etc.) are effectively exploited [4]–[6], [13], [15], [31], [34]. Nesterov's accelerated methods have also been widely adopted in training deep neural networks and significantly increased their performance [14].

In contrast to smooth objective cases, there is no specific difficulty preventing almost all the subgradient algorithms to achieve optimal rates of $O((1/\sqrt{t}))$. In spite of various progress in convergence rates (for instance, see [2] and references therein), several fundamental questions such as the attainable convergence rates still remain unsolved. Almost all the optimal convergence rates in nonsmooth cases are only obtained by averaging all past iterates. However, as far as the strongly convex objective is concerned, simply averaging all past iterates is not enough, as it only corresponds to the suboptimality rate of $O((\log t/t))$ as opposed to the expected optimal rate of $O((1/t))$ [25]. Moreover, an example given in [24] shows that the convergence rate of stochastic gradient descent (SGD) with the averaged output is exactly

Manuscript received February 28, 2019; revised June 29, 2019 and July 30, 2019; accepted August 2, 2019. This work was supported in part by NSFC under Grant 61673394 and in part by the National Key Research and Development Program of China under Grant 2016QY03D0501. (*Corresponding author: Qing Tao.*)
W. Tao and Z. Pan are with the Command and Control Engineering College, Army Engineering University of PLA, Nanjing 210007, China (e-mail: wtao_plaust@163.com).
G. Wu is with the Institute of Automation, Chinese Academy of Sciences, China (e-mail: gaowei.wu@ia.ac.cn).
Q. Tao is with the Institute of Automation, Chinese Academy of Sciences, China, and also with the Army Academy Of Artillery and Air Defense, Hefei 230031, China (e-mail: taoqing@gmail.com).
Color versions of one or more of the figures in this article are available online at http://ieeexplore.ieee.org.
Digital Object Identifier 10.1109/TNNLS.2019.2933452

$\Omega((\log t/t))$ in nonsmooth and strongly convex cases. Due to these facts, an open question was posed in [26] whether the averaging scheme is required to reach optimal convergence.

During the last few years, there have been many interesting developments to make SGD optimal. One important idea is based on appropriately modifying the averaging scheme. By averaging only a suffix of the iterates [24] or by weighted averaging of all past solutions [17], the logarithmic factor within the suboptimality rate can be removed. While these theoretical results show that we should use improved averaging schemes, empirically, people have mostly been using only the last iteration as the final solution [24], [25]. Intuitively, an individual output directly decided by the step of gradient operation potentially has the advantage of readily enforcing the regularization structure [6]. These facts suggest that we should study convergence rates of the last individual iterate theoretically, which is also accordant with the traditional optimization theory. In 2013, the first finite-sample bounds on individual iterates of SGD were established [27]. However, it only has a suboptimal $O((\log t/\sqrt{t}))$ rate for general nonsmooth convex objective functions, and a suboptimal $O((\log t/t))$ rate in nonsmooth but strongly convex cases. So far, there still exists a gap between practical performance and theoretical analysis. In this article, we will focus on the convergence of the last individual iterate of PSG in nonsmooth cases, which is referred to as *individual convergence* for simplicity.

Several significant efforts have been paid to close the aforementioned gap by using other first-order gradient algorithms. In particular, Chen et al. [32] improved the regularized dual averaging method (RDA) to be self-adaptive and obtained uniformly optimal individual convergence for nonsmooth convex functions [6]. However, their optimal RDA is remarkably different from the regular RDA in computation, that is, the former requires two steps of subgradient computation at each iteration, rather than only one step of subgradient operation used in the latter. As a result, the subgradient operation becomes less intuitive. Recently, by incorporating the averaging scheme into DA directly, Nesterov et al. [22] succeeded in deriving an optimal individual convergence rate of $O((1/\sqrt{t}))$ for nonsmooth convex problems. However, the strongly convex case is not discussed. It should be mentioned that the derivations of optimal individual convergence so far only focus on the DA-like methods. Motivated by the averaging step in quasi-monotone DA [22], we recently presented a primal averaging (PA) strategy for PSG [29], in which the subgradient evaluation is imposed on the average of all past iterates. The PA strategy can accelerate the individual convergence of PSG to be optimal for nonsmooth convex problems. In light of this consideration, an optimal rate of individual convergence for strongly convex problems can also be obtained by further modifying the PA step.

Inspired by the great success of Nesterov's extrapolation step in smooth optimization, we will study its convergence rate in nonsmooth optimization. In particular, we discover that Nesterov's extrapolation strategy is capable of accelerating the individual convergence of fundamental PSG to be optimal for general convex problems without the additional assumption of smoothness. Further, a simple modification of the gradient operation suffices to achieve optimal individual convergence for strongly convex problems. These results indicate that the averaging scheme to attain optimal rates can be unnecessary. Although whether an optimal individual rate of SGD can be attained is still unknown, we can simply achieve optimal individual convergence with the help of Nesterov's extrapolation, which might be regarded as making an interesting step toward the open question about SGD posed by Shamir [26]. No matter there is a smoothness assumption or not, Nesterov's extrapolation consistently has the strength to provide theoretical guarantee of optimal convergence. The derived algorithms in this article can serve as an alternative to the fundamental SGD especially in the machine learning community, where an individual output is needed to guarantee the regularization structure while keeping an optimal rate of convergence. Moreover, the novelty of convergence analysis presented in this article reveals more insights about the similarities and differences between smooth and nonsmooth optimization methods, and then help us understand how to correctly and provably extend smooth optimization methods to nonsmooth tasks.

It should be noted that the effect of Nesterov's accelerated strategy on nonsmooth problems has been investigated in [16] and [8]. Specifically, a modified mirror descent (MD) algorithm [called accelerated stochastic approximation (AC-SA)] was proven to attain optimal individual convergence for general convex problems [16], in which the same stepsize policy as the second accelerated method [30] is employed. Unfortunately, as stated in [7], AC-SA assumes a priori knowledge of the performed number of iterations and use stepsizes based on this number. On the other hand, a fast gradient method (FGM) developed originally for smooth optimization in [21] was proven to be optimal for nonsmooth problems in terms of the individual convergence [8]. However, FGM suffers from the same drawback as ORDA [6] in needing two steps of gradient evaluation per iteration. Fortunately, our algorithms avoid the disadvantages of both AC-SA and FGM while keeping optimal individual convergence. The key to success lies in our direct use of Nesterov's extrapolation strategy and suitable selection of the time-varying stepsize parameters. Moreover, the optimal individual convergence for challenging strongly convex problems can be further derived.

Based upon our convergence analysis for black-box problems, the derived algorithms are extended to regularized and stochastic settings for large scale machine learning tasks. Unlike the quasi-monotone algorithm [22] or PA-PSG [29], the subgradientlike operation in our method follows the extrapolation evaluation, which brings significant benefits in keeping the regularization structure. Our method is applicable as an efficient tool for solving large-scale $l_1$-regularized hinge-loss learning problems. It not only guarantees the best possible rate of individual convergence but also achieves better sparsity than the averaged solution. Several experiments confirm the correctness of our convergence analysis and illustrate the performance of our algorithms in keeping sparsity.

The rest of this article is organized as follows. Section II reviews the success of Nesterov's extrapolation in smooth optimization. Section III reviews some related first-order subgradient methods in nonsmooth cases. In Sections IV and V, optimal individual convergence of PSG with Nesterov's extrapolation for general convex and strongly convex problems are proved, respectively. The optimal individual convergence results are extended to regularized and stochastic settings in Section VI. Comparison experiments are conducted in Section V, and the concluding remarks follow in the last section.

## II. NESTEROV'S EXTRAPOLATION IN SMOOTH OPTIMIZATION

In this section, we briefly review the strength of Nesterov's extrapolation in smooth optimization. To this end, we assume that the objective function $f$ in (1) is smooth, that is, there exists a constant $L \geq 0$ such that

$$f(\mathbf{w}) \leq f(\mathbf{u}) + \langle \nabla f(\mathbf{u}), \mathbf{w} - \mathbf{u} \rangle + \frac{L}{2}\|\mathbf{w} - \mathbf{u}\|^2 \quad (2)$$

$\forall \mathbf{u}, \mathbf{w} \in \mathbf{Q}$.

The iteration of PSG is

$$\mathbf{w}_{t+1} = P[\mathbf{w}_t - a_t \nabla f(\mathbf{w}_t)] \quad (3)$$

or equivalently

$$\mathbf{w}_{t+1} = \arg\min_{\mathbf{w} \in \mathbf{Q}} \left\{ a_t \langle \nabla f(\mathbf{w}_t), \mathbf{w} \rangle + \frac{1}{2}\|\mathbf{w} - \mathbf{w}_t\|^2 \right\} \quad (4)$$

where $P$ is the projection operator on $\mathbf{Q}$ [2], $a_t$ is the parameter about step-size, and $\nabla f(\mathbf{w}_t)$ is the gradient of $f$ at $\mathbf{w}_t$. Let $\mathbf{w}^*$ denote an optimal solution of problem (1).

Specifically when $a_t = (1/L)$, it holds that [30]

$$f(\mathbf{w}_t) - f(\mathbf{w}^*) \leq \frac{L}{2t}\|\mathbf{w} - \mathbf{w}_0\|^2. \quad (5)$$

In a series of work (see also [30]), Nesterov proposed three methods for solving smooth problems (1) with constraints that, at each iteration, use either one or two steps of gradient operation together with extrapolation to accelerate convergence. In particular, the first accelerated method can be formulated as

$$\begin{cases} \mathbf{y}_t = \mathbf{w}_t + \theta_t(\theta_{t-1}^{-1} - 1)(\mathbf{w}_t - \mathbf{w}_{t-1}) \\ \mathbf{w}_{t+1} = \arg\min_{\mathbf{w} \in \mathbf{Q}} \left\{ \langle \nabla f(\mathbf{y}_t), \mathbf{w} \rangle + \frac{L}{2}\|\mathbf{w} - \mathbf{y}_t\|^2 \right\} \\ \text{Choose } \theta_{t+1} \in (0, 1] \text{ satisfying} \\ \frac{1 - \theta_{t+1}}{\theta_{t+1}^2} \leq \frac{1}{\theta_t^2} \end{cases} \quad (6)$$

where $\mathbf{w}_0 = \mathbf{w}_{-1} \in \mathbb{R}^N$ and $\theta_0 = \theta_{-1} = 1$.

Note the stepsize choice of $\theta_{t+1}$ in (6) is given by Tseng [30]. It can be regarded as a generalization of the original rule in [19]. According to this generalized rule, one choice of $\theta_{t+1}$ is

$$\theta_{t+1} = \frac{2}{t+2}$$

which is also employed in SAGE [13]. Another choice of $\theta_{t+1}$ is

$$\theta_{t+1} = \frac{\sqrt{\theta_t^4 + 4\theta_t^2} - \theta_t^2}{2}$$

which is also used in FISTA [1].

From

$$\frac{\theta_{t+1}}{1 - \theta_{t+1}^2} \leq \frac{1}{\theta_t^2}$$

an inductive argument shows that $\theta_{t+1} \leq (2/t+2)$. As $t \to \infty$, $(\theta_t/\theta_{t-1}) \leq (1 - \theta_t)^{1/2} \to 1$. Thus $\mathbf{y}_t$, which the gradient operation is imposed on, is a extrapolation from $\mathbf{w}_{t-1}$ to $\mathbf{w}_t$. The stepsize rule

$$\mathbf{y}_t = \mathbf{w}_t + \theta_t(\theta_{t-1}^{-1} - 1)(\mathbf{w}_t - \mathbf{w}_{t-1})$$

in (6) is referred to as *Nesterov's extrapolation* in this article.

The second and third accelerated methods are also described in [30]. Despite different extrapolation steps are used, all these three methods are equivalent [30] when the concerned Bregman divergence is Euclidian. Without loss of generality, we mainly focus on the first accelerated method (6).

Let $\{\mathbf{w}_t\}_{t=1}^{\infty}$ be generated by the first accelerated method (6). For any $\mathbf{w} \in \mathbb{R}^N$, it holds that [30]

$$f(\mathbf{w}_t) - f(\mathbf{w}) \leq \frac{\theta_{t-1}^2 L}{2}\|\mathbf{w} - \mathbf{w}_0\|^2.$$

This bound means that the convergence rate of the accelerated method (6) is $O((1/t^2))$, which clearly exhibits the strength of Nesterov's extrapolation in obtaining optimal convergence. So far, stochastic Nesterov's accelerated methods have been employed to solve large-scale regularized learning problems [6], [13], [31]. In addition to accelerating the convergence in the cases of smooth loss but nonsmooth regularizer, each accelerated method obtains an optimal rate of convergence in terms of the individual output, which is highly expected in nonsmooth optimization.

## III. PSG IN NONSMOOTH OPTIMIZATION

This section reviews several significant results about the convergence of PSG. We now assume that the objective function $f$ in (1) is nonsmooth.

Let $A_t = \sum_{k=1}^{t} a_k$ and $\{\mathbf{w}_t\}_{t=1}^{\infty}$ be generated by PSG (3), it holds that [22]

$$\frac{1}{A_t}\sum_{k=1}^{t} a_k f(\mathbf{w}_k) - f(\mathbf{w}^*)$$
$$\leq \frac{1}{A_t}\left[\frac{1}{2}\|\mathbf{w}_0 - \mathbf{w}^*\|^2 + \sum_{k=1}^{t} \frac{a_k^2}{2}\|\nabla f(\mathbf{w}_k)\|^2\right]. \quad (7)$$

To get concrete convergence rates, we require

*Assumption 1:* Let $\nabla f(\mathbf{w})$ denote any subgradient of $f$ at $\mathbf{w}$. Assume that there exists a number $M > 0$ such that

$$\|\nabla f(\mathbf{w})\| \leq M \quad \forall \mathbf{w} \in \mathbf{Q}.$$

With suitable $a_t$, an optimal rate of convergence can be derived for nonsmooth problems [22], that is,

$$f\left(\frac{1}{A_t}\sum_{k=1}^{t} a_k \mathbf{w}_k\right) - f(\mathbf{w}^*) \leq O\left(\frac{1}{\sqrt{t}}\right).$$

Let $\hat{\mathbf{g}}_t$ be an unbiased estimate of subgradient of $f$ at $\mathbf{w}_t$. The PSG (3) can be simply extended to stochastic settings, in which the key operation is

$$\mathbf{w}_{t+1} = P(\mathbf{w}_t - a_t \hat{\mathbf{g}}_t). \tag{8}$$

To get the convergence rates in stochastic settings, we require

*Assumption 2:* Let $\hat{\mathbf{g}}$ be an unbiased estimate of subgradient of $f$ at $\mathbf{w}$. Assume that

$$\mathbb{E}\|\hat{\mathbf{g}} - \nabla f(\mathbf{w})\|^2 \leq \sigma^2 \quad \forall \mathbf{w} \in \mathbf{Q}. \tag{9}$$

Based on the regret bounds in [35], it is easy to know that the stochastic PSG (8) has

$$\mathbb{E}\left[f\left(\frac{1}{t}\sum_{k=1}^{t}\mathbf{w}_k\right)\right] - f(\mathbf{w}^*) \leq O\left(\frac{1}{\sqrt{t}}\right).$$

While an optimal rate of convergence is easily attained, it is only in terms of the averaged output $\frac{1}{t}\sum_{k=1}^{t}\mathbf{w}_k$. When $f$ is strongly convex, even the regret bounds in [11] and [12] are employed, we can only get

$$\mathbb{E}\left[f\left(\frac{1}{t}\sum_{k=1}^{t}\mathbf{w}_k\right)\right] - f(\mathbf{w}^*) \leq O\left(\frac{\log t}{t}\right).$$

This fact indicates that even the averaged output $(1/t)\sum_{k=1}^{t}\mathbf{w}_k$ can not lead to an optimal rate due to a logarithmic factor.

To obtain individual convergence without changing the original algorithms, a general technique was presented in [27] to reduce results on averages of iterates to convergence on individual iterates. As a result, they have gotten so far the best rates for stochastic PSG in terms of the individual convergence (3), that is,

$$\mathbb{E}[f(\mathbf{w}_t)] - f(\mathbf{w}^*)] \leq O\left(\frac{\log t}{\sqrt{t}}\right)$$

for nonsmooth convex problems and

$$\mathbb{E}[f(\mathbf{w}_t)] - f(\mathbf{w}^*)] \leq O\left(\frac{\log t}{t}\right)$$

for strongly convex problems.

To get first-order gradient algorithms with the expected optimal individual convergence, there are some important works on Nesterov's DA. Typical examples include optimal RDA [6] and quasi-monotone DA [22]. Specifically, the quasi-monotone subgradient method is formulated as

$$\begin{cases} \mathbf{w}_t^+ = \arg\min_{\mathbf{w}\in\mathbf{Q}} \left\{ \sum_{k=1}^{t}\langle a_k\nabla f(\mathbf{w}_k), \mathbf{w}\rangle + \gamma_t d(\mathbf{w}) \right\} \\ \mathbf{w}_{t+1} = \frac{A_t}{A_{t+1}}\mathbf{w}_t + \frac{a_{t+1}}{A_{t+1}}\mathbf{w}_t^+. \end{cases} \tag{10}$$

Note that the additional averaging parameters in (10) is a specified interpolation step. With suitable $a_t$ and $\gamma_t$, it holds that [22]

$$f(\mathbf{w}_t) - f(\mathbf{w}^*) \leq O\left(\frac{1}{\sqrt{t}}\right).$$

In order to make PSG optimal in terms of the individual convergence for general convex problems, a stochastic PSG with PA operation (PA-PSG) is given in [29], that is,

$$\begin{cases} \mathbf{w}_t^+ = P\left(\mathbf{w}_{t-1}^+ - \frac{a_t}{\gamma_t}\hat{\mathbf{g}}_t\right) \\ \mathbf{w}_{t+1} = \frac{A_t}{A_{t+1}}\mathbf{w}_t + \frac{a_{t+1}}{A_{t+1}}\mathbf{w}_t^+ \end{cases} \tag{11}$$

where $\hat{\mathbf{g}}_t$ is an unbiased estimate of subgradient of $f$ at $\mathbf{w}_t$. It is easy to find that the gradient operation in (11) is imposed on $f(\mathbf{w}_t)$, in which $\mathbf{w}_t$ in fact is a weighted average of all past iterates.

When the objective in (1) is $\mu$-strongly convex, stochastic PA-PSG [29] is modified as

$$\begin{cases} \mathbf{w}_t^+ = P\left[\frac{\gamma_t}{\gamma_t + a_t\mu}\mathbf{w}_{t-1}^+ - \frac{a_t}{\gamma_t + a_t\mu}(\hat{\mathbf{g}}_t - \mu\mathbf{w}_t)\right] \\ \mathbf{w}_{t+1} = \frac{A_t}{A_{t+1}}\mathbf{w}_t + \frac{a_{t+1}}{A_{t+1}}\mathbf{w}_t^+. \end{cases} \tag{12}$$

With suitable $a_t$ and $\gamma_t$ [29], PA-PSG (12) has

$$\mathbb{E}f(\mathbf{w}_t) - f(\mathbf{w}^*) \leq O\left(\frac{1}{t}\right).$$

However, when dealing with regularized learning problems, we don't know whether the regularization structure can be guaranteed by employing (11). One of the main purposes of this article is to make PSG optimal in terms of the individual convergence while keeping the regularization structure. The main idea is to employ Nesterov's extrapolation in nonsmooth optimization.

## IV. NESTEROV'S EXTRAPOLATION IN NONSMOOTH OPTIMIZATION

In this section, we study the strength of Nesterov's extrapolation in improving the individual convergence of basic PSG. Naturally, we consider

$$\begin{cases} \mathbf{y}_t = \mathbf{w}_t + \theta_t(\theta_{t-1}^{-1} - 1)(\mathbf{w}_t - \mathbf{w}_{t-1}) \\ \mathbf{w}_{t+1} = P[\mathbf{y}_t - a_t\nabla f(\mathbf{y}_t)] \\ \text{Choose } \theta_{t+1} \in (0, 1] \text{ satisfying} \\ \frac{1 - \theta_{t+1}}{\theta_{t+1}^2} \leq \frac{1}{\theta_t^2} \end{cases} \tag{13}$$

In contrast to regular PSG (3) and PA-PSG (11), we call (13) a PSG with Nesterov's extrapolation or Nesterov's PSG. The only difference between (6) and (13) lies in the selection of step size. It looks like that the convergence analysis will be completely similar to that of accelerated methods for smooth optimization in [30]. However, the convergence analysis in smooth optimization [30] heavily depends on the property

$$f(\mathbf{w}) \leq f(\mathbf{u}) + \langle\nabla f(\mathbf{u}), \mathbf{w} - \mathbf{u}\rangle + \frac{L}{2}\|\mathbf{w} - \mathbf{u}\|^2$$

which no longer holds for nonsmooth functions. To get convergence rates without smooth assumption, the key issue is how to deal with some terms caused by the time-varying stepsize and some additional terms caused by the nonsmooth objective function.

To conduct a convergence analysis, a few lemmas are required. We only give the main theorem here. All the lemmas (Lemmas 1, 2, 3, and 4) with proof details are shown in the Appendix.

*Theorem 1:* Let $\{\mathbf{w}_t\}_{t=1}^{\infty}$ be generated by (13). Let $\theta_t = (2/t+1)$ and $a_t = (1/(t+1)(t+1)^{1/2})$. For any $\mathbf{w} \in \mathbf{Q}$, it holds

$$f(\mathbf{w}_t) - f(\mathbf{w}^*) \leq O\left(\frac{1}{\sqrt{t}}\right).$$

Without modifying the regular stepsize rule, PSG only has a suboptimal $O((\log t/\sqrt{t}))$ individual rate for nonsmooth convex objective functions [27]. By simply replacing the original diminishing stepsize rule with the Nesterov's extrapolation step originally for smooth optimization, we prove that Nesterov's PSG (13) has optimal individual convergence for nonsmooth problems. Naturally, its stochastic version can be obtained by directly substituting the gradient $\nabla f(\mathbf{y}_t)$ in (13) with its unbiased estimate $\hat{\mathbf{g}}_t$. Such a substitution will not affect its optimal individual convergence. In fact, it is not difficult to find that Lemma 1 becomes

$$a_t[f(\mathbf{y}_t) - f(\mathbf{w})]$$
$$\leq \frac{1}{2}\|\mathbf{w} - \mathbf{y}_t\|^2 - \frac{1}{2}\|\mathbf{w} - \mathbf{w}_{t+1}\|^2 - \frac{1}{2}\|\mathbf{y}_t - \mathbf{w}_{t+1}\|^2$$
$$+ a_t\langle \nabla f(\mathbf{y}_t) - \hat{\mathbf{g}}_t, \mathbf{y}_t - \mathbf{w}\rangle + a_t\langle \hat{\mathbf{g}}_t, \mathbf{y}_t - \mathbf{w}_{t+1}\rangle.$$

By using Assumptions 1 and 2

$$a_t\langle \hat{\mathbf{g}}_t, \mathbf{y}_t - \mathbf{w}_{t+1}\rangle - \frac{1}{2}\|\mathbf{y}_t - \mathbf{w}_{t+1}\|^2 \leq \frac{1}{2}a_t^2(M^2 + \sigma^2)$$

Note (see [7, proof of Th. 1, p. 9] and [16, eq. (58), p. 21])

$$\mathbb{E}\langle \nabla f(\mathbf{y}_t) - \hat{\mathbf{g}}_t, \mathbf{y}_t - \mathbf{w}\rangle = 0.$$

Now, it becomes clear that similar bounds like that in Lemma 2 hold for stochastic Nesterov's PSG. Using the same deduction as that in Lemmas 3 and 4, we can prove that Nesterov's PSG (13) in the stochastic setting achieves an optimal individual convergence rate in expectation for nonsmooth convex problems.

It should be mentioned that Nesterov's accelerated methods have been applied in nonsmooth optimizations [8], [16]. To make a careful comparison, we give some remarks here.

1) The key operation of AC-SA algorithm in [16] can be formulated as

$$\begin{cases} \mathbf{w}_t^{md} = \beta_t^{-1}\mathbf{w}_t + (1 - \beta_t^{-1})\mathbf{w}_t^{ag} \\ \mathbf{w}_{t+1} = P[\mathbf{w}_t^{md} - \gamma_t \nabla f(\mathbf{w}_t^{md})] \\ \mathbf{w}_{t+1}^{ag} = \beta_t^{-1}\mathbf{w}_{t+1} + (1 - \beta_t^{-1})\mathbf{w}_t^{ag} \end{cases}$$

which is stated as the second accelerated method in [30] for smooth optimization. It was proven to attain an optimal individual convergence rate for nonsmooth optimization but requires $\beta_t$ to depend on the performed number of iterates [16]. As stated in [7], this assumption is not desirable when we want to run a method for a given time and not for a given number of iterates.

2) FGM was originally developed for smooth optimization [21]. As a byproduct, it was proven to be optimal for nonsmooth problems in terms of the individual convergence [8]. However, at each iteration, there are two steps of subgradient operation in FGM, which results in the same disadvantage as that in ORDA [6]. In [30], FGM was improved to have only one projection instead of two, which is also called the third accelerated method. But this accelerated method is only limited to smooth optimization problems [30].

3) In contrast to FGM, our method uses time-varying stepsizes that does not need to know the performed number of iterates in advance. Based on the equivalence between the three accelerated methods [30], we can say that our method avoids the disadvantages of both AC-SA and FGM in coping with nonsmooth optimization. Moreover, its optimal individual convergence in strongly convex cases can be further achieved, which will be discussed in the next section.

## V. NESTEROV'S EXTRAPOLATION IN STRONGLY CONVEX CASES

In this section, we will discuss the strongly convex cases, that is, there exists a real number $\mu > 0$ such that

$$f(\mathbf{w}) \geq f(\mathbf{u}) + \langle \nabla f(\mathbf{u}), \mathbf{w} - \mathbf{u}\rangle + \frac{\mu}{2}\|\mathbf{w} - \mathbf{u}\|^2$$

or equivalently [2]

$$f(\theta\mathbf{w} + (1-\theta)\mathbf{u}) \leq \theta f(\mathbf{w}) + (1-\theta)f(\mathbf{u}) - \frac{\mu}{2}\theta(1-\theta)\|\mathbf{w} - \mathbf{u}\|^2 \quad (14)$$

$\forall \mathbf{u}, \mathbf{w} \in \mathbb{R}^N$.

To utilize the strong convexity of $f$, we modify (15) as

$$\begin{cases} \mathbf{y}_t = \mathbf{w}_t + \theta_t(\theta_{t-1}^{-1} - 1)(\mathbf{w}_t - \mathbf{w}_{t-1}) \\ \mathbf{w}_{t+1} = P\left[\frac{\theta_t}{\theta_t + a_t\mu}\mathbf{y}_t + \frac{a_t\mu}{\theta_t + a_t\mu}\mathbf{w}_t - \frac{a_t\theta_t}{\theta_t + a_t\mu}\nabla f(\mathbf{y}_t)\right] \\ \text{Choose } \theta_{t+1} \in (0, 1] \text{ satisfying} \\ \frac{1 - \theta_{t+1} + \theta_{t+1}^2}{\theta_{t+1}^2} \leq \frac{1}{\theta_t^2}. \end{cases} \quad (15)$$

It is easy to find that algorithm (15) is special case of (13) when $\mu = 0$. Without loss of generality, we will assume $\mu \neq 0$ in the following. Obviously, in strongly convex cases, the key to derive optimal individual convergence lies in how to get stronger results from the analysis in Section IV. Specifically, we give Lemmas 6 and 7 in the Appendix. Based on Lemmas 6 and 7, we have

*Theorem 1:* Let $\{\mathbf{w}_t\}_{t=1}^{\infty}$ be generated by (15). Let $a_t = (3/\mu t^2)$ and

$$\theta_t = \begin{cases} 1, & \text{if } t \leq 7 \\ \dfrac{3}{t+1}, & \text{otherwise} \end{cases}$$

it holds

$$f(\mathbf{w}_t) - f(\mathbf{w}^*) \leq O\left(\frac{1}{t}\right). \quad (16)$$

Without modifying the regular gradient operation and stepsize rule, PSG only has a suboptimal $O((\log t/t))$ individual rate in nonsmooth but strongly convex cases [27]. By imposing on the strong convexity coming from the objective function and Nesterov's extrapolation step originally for smooth optimizations, our PSG (15) achieves an optimal rate of individual convergence for strongly convex problems.

Like that in general convex cases, PSG with Nesterov's extrapolation (15) can also be extended to stochastic settings by substituting the gradient $\nabla f(\mathbf{y}_t)$ in (15) with its unbiased estimate $\hat{\mathbf{g}}_t$. Under Assumptions 1 and 2, we can obtain a similar optimal convergence bound of Theorem 1 in expectation, that is, by choosing identical $\theta_t$ and $a_t$, we can derive

$$\mathbb{E}[f(\mathbf{w}_t) - f(\mathbf{w}^*)] \leq O\left(\frac{1}{t}\right).$$

## VI. EXTENSION TO REGULARIZED LEARNING

In this section, we study the regularized learning problems, that is,

$$\min F(\mathbf{w}) = r(\mathbf{w}) + f(\mathbf{w}), \quad \text{s.t.} \quad \mathbf{w} \in \mathbf{Q} \quad (17)$$

where $\mathbf{Q} \subseteq \mathbb{R}^N$ is a closed convex set, and $r(\mathbf{w}) : \mathbb{R}^N \to \mathbb{R}$ is a simple convex function and $f(\mathbf{w}) = (1/m) \sum_{i=1}^m f_i(\mathbf{w})$. Each $f_i(\mathbf{w}) : \mathbb{R}^N \to \mathbb{R}$ is a nonsmooth convex loss function.

Like the regularizing technique in dealing with accelerated PSG [30], MD [10] and DA [31], the regularization term should not be linearized. Naturally, our stochastic regularized subgradient (SRSG) with Nesterov's extrapolation takes the form

$$\begin{cases} \mathbf{y}_t = \mathbf{w}_t + \theta_t(\theta_{t-1}^{-1} - 1)(\mathbf{w}_t - \mathbf{w}_{t-1}) \\ \mathbf{w}_{t+1} = \arg\min_{\mathbf{w} \in \mathbf{Q}} \left\{ a_t \langle \hat{\mathbf{g}}_t, \mathbf{w} \rangle + a_t r(\mathbf{w}) + \frac{1}{2} \|\mathbf{w} - \mathbf{y}_t\|^2 \right\} \\ \text{Choose } \theta_{t+1} \in (0, 1] \text{ satisfying} \\ \frac{1 - \theta_{t+1}}{\theta_{t+1}^2} \leq \frac{1}{\theta_t^2} \end{cases} \quad (18)$$

where $\hat{\mathbf{g}}_t$ is an unbiased estimate of subgradient of $f$ at $\mathbf{y}_t$.

Similarly, when $f$ is $\mu$-strongly convex, the SRSG with Nesterov's extrapolation can be formulated as

$$\begin{cases} \mathbf{y}_t = \mathbf{w}_t + \theta_t(\theta_{t-1}^{-1} - 1)(\mathbf{w}_t - \mathbf{w}_{t-1}) \\ \mathbf{w}_{t+1} = \arg\min_{\mathbf{w} \in \mathbf{Q}} \left\{ a_t \langle \hat{\mathbf{g}}_t, \mathbf{w} \rangle + a_t r(\mathbf{w}) + \frac{1}{2} \|\mathbf{w} - \mathbf{y}_t\|^2 \right. \\ \left. \qquad\qquad + \frac{a_t}{2\theta_t} \mu \|\mathbf{w} - \mathbf{w}_t\|^2 \right\} \\ \text{Choose } \theta_{t+1} \in (0, 1] \text{ satisfying} \\ \frac{1 - \theta_{t+1} + \theta_{t+1}^2}{\theta_{t+1}^2} \leq \frac{1}{\theta_t^2}. \end{cases} \quad (19)$$

To get concrete convergence rates, we require

*Assumption 3:* Assume that there exists a number $M > 0$ such that

$$\|\nabla r(\mathbf{w}) + \nabla f(\mathbf{w})\| \leq M \quad \forall \mathbf{w} \in \mathbf{Q}.$$

*Assumption 4:* Let $\hat{\mathbf{g}}$ be an unbiased estimate of subgradient of $f$ at $\mathbf{w}$. Assume that

$$\mathbb{E}\|\nabla r(\mathbf{w}) + \hat{\mathbf{g}} - \nabla F(\mathbf{w})\|^2 \leq \sigma^2 \quad \forall \mathbf{w} \in \mathbf{Q}. \quad (20)$$

The regularized reformulation will not affect the optimal individual convergence we have derived in Theorems 1 and 2. In fact, it is not difficult to find that Lemma 1 becomes

$$\begin{aligned} a_t[r(\mathbf{y}_t) &+ f(\mathbf{y}_t) - r(\mathbf{w}) - f(\mathbf{w})] \\ &\leq \frac{1}{2} \|\mathbf{w} - \mathbf{y}_t\|^2 - \frac{1}{2} \|\mathbf{w} - \mathbf{w}_{t+1}\|^2 - \frac{1}{2} \|\mathbf{y}_t - \mathbf{w}_{t+1}\|^2 \\ &\quad + a_t \langle \nabla f(\mathbf{y}_t) - \hat{\mathbf{g}}_t, \mathbf{y}_t - \mathbf{w} \rangle + a_t \langle \hat{\mathbf{g}}_t, \mathbf{y}_t - \mathbf{w}_{t+1} \rangle \\ &\quad + a_t[r(\mathbf{y}_t) - r(\mathbf{w}_{t+1})] \end{aligned}$$

where $\hat{\mathbf{g}}_t$ is an unbiased estimate of subgradient of $f$ at $\mathbf{y}_t$. Note

$$r(\mathbf{y}_t) - r(\mathbf{w}_{t+1}) \leq \langle \nabla r(\mathbf{y}_t), \mathbf{y}_t - \mathbf{w}_{t+1} \rangle.$$

Then

$$\begin{aligned} a_t[r(\mathbf{y}_t) &+ f(\mathbf{y}_t) - r(\mathbf{w}) - f(\mathbf{w})] \\ &\leq \frac{1}{2} \|\mathbf{w} - \mathbf{y}_t\|^2 - \frac{1}{2} \|\mathbf{w} - \mathbf{w}_{t+1}\|^2 - \frac{1}{2} \|\mathbf{y}_t - \mathbf{w}_{t+1}\|^2 \\ &\quad + a_t \langle \nabla f(\mathbf{y}_t) - \hat{\mathbf{g}}_t, \mathbf{y}_t - \mathbf{w} \rangle + a_t \langle \nabla r(\mathbf{y}_t) + \hat{\mathbf{g}}_t, \mathbf{y}_t - \mathbf{w}_{t+1} \rangle. \end{aligned}$$

Under Assumptions 3 and 4, similar bounds as in Lemmas 2 and 7 hold for SRSG with Nesterov's extrapolation (18) and (19). Using the same deduction as that in Sections III and IV, we can prove that SRSG with Nesterov's extrapolation (18) and (19) attains optimal individual convergence in expectation for nonsmooth convex and strongly convex problems respectively, that is, we have

*Theorem 1:* Let $\mathbf{w}_0 \in \mathbf{Q}$ be an initial point and $\{\mathbf{w}_t\}_{t=1}^{\infty}$ be generated by SRSG with Nesterov's extrapolation (18). Let $\theta_t = (2/t + 1)$ and $a_t = (1/(t + 1)(t + 1)^{1/2})$. We have

$$\mathbb{E}[F(\mathbf{w}_t) - F(\mathbf{w}^*)] \leq O\left(\frac{1}{\sqrt{t}}\right)$$

where $\mathbf{w}^*$ is an optimal solution of problem (17).

*Theorem 2:* Assume $\mu > 0$ and that $f$ is $\mu$-strongly convex. Let $\mathbf{w}_0 \in \mathbf{Q}$ be an initial point and $\{\mathbf{w}_t\}_{t=1}^{\infty}$ be generated by SRSG with Nesterov's extrapolation (19). Let $a_t = (3/\mu t^2)$ and

$$\theta_t = \begin{cases} 1, & \text{if } t \leq 7 \\ \dfrac{3}{t+1}, & \text{otherwise.} \end{cases}$$

We have

$$\mathbb{E}[F(\mathbf{w}_t) - F(\mathbf{w}^*)] \leq O\left(\frac{1}{t}\right). \quad (21)$$

In [29], PA-like algorithms are extended to solve regularized nonsmooth loss optimization problems in stochastic settings. Specifically, PA-PSG (11) is reformulated as

$$\begin{cases} \mathbf{w}_t^+ = \arg\min_{\mathbf{w} \in Q} \left\{ a_t \langle \hat{\mathbf{g}}_t, \mathbf{w} \rangle + \gamma_t B(\mathbf{w}, \mathbf{w}_{t-1}^+) + a_t r(\mathbf{w}) \right\} \\ \mathbf{w}_{t+1} = \dfrac{A_t}{A_{t+1}} \mathbf{w}_t + \dfrac{a_{t+1}}{A_{t+1}} \mathbf{w}_t^+ \end{cases} \quad (22)$$

where $B$ is the Bregman divergence and $\hat{\mathbf{g}}_t$ is an unbiased estimate of subgradient of $f$ at $\mathbf{w}_t$. When solving $l_1$-regularized learning problems, each $\mathbf{w}_t^+$ in (22) is derived by solving the optimization subproblem in closed-form and thus has better sparsity than the averaged solution. Unfortunately, $\mathbf{w}_{t+1} = (A_t/A_{t+1})\mathbf{w}_t + (a_{t+1}/A_{t+1})\mathbf{w}_t^+$ can be reformulated as

$$\mathbf{w}_{t+1} = \frac{1}{A_t}\sum_{k=1}^{t} a_k \mathbf{w}_k^+.$$

TABLE I
REAL DATA SETS WHERE THE SCALE DESCRIBES
THE SIZE OF A TRAINING SET

| DATA-SET | DIMENSION | SCALE |
|---|---|---|
| A9A | 123 | 32,561 |
| RCV1 | 47,236 | 20,242 |
| SIDO | 4,932 | 12,678 |
| COVTYPE | 54 | 522,911 |

This means the individual solution $\mathbf{w}_{t+1}$ in (22) is a linear combination of $\mathbf{w}_k^+$ ($k = 1, 2, \ldots, t$). As a result, the structure especially imposed on by the $l_1$-regularizer may not be guaranteed.

In contrast to PA-PSG (22), the individual solution of SRSG with Nesterov's extrapolation is directly generated by the closed-form solution of the optimization subproblem. The sparsity decided by the $l_1$-regularizer is then effectively kept.

## VII. EXPERIMENTS

In this section, we will conduct experiments to verify our theoretical analysis and illustrate the performance of our algorithms in keeping the sparsity. To clearly illustrate its advantage over other strategies, we focus on the comparison experiments on PSG with different stepsize rules in terms of convergence and sparsity (i.e., the percentage of nonzero entries, [34]). Four benchmark data sets in Table I are considered, where one data set is available at http://www.causality.inf.ethz.ch/data/SIDO.html and the other three data sets are available at http://www.csie.ntu.edu.tw/∼cjlin/libsvmtools/datasets/.

Let $\mathcal{S} = \{(\mathbf{x}_1, y_1), \ldots, (\mathbf{x}_m, y_m)\}$ be a training set, where $y_i \in Y = \{-1, 1\}$ is the label of $\mathbf{x}_i$. Let $f_i(\mathbf{w}) = l(\mathbf{w}, \mathbf{x}_i, y_i)$ be the convex and nonsmooth loss caused by $(\mathbf{x}_i, y_i)$. At the $t$-th step of the stochastic algorithms, we assume

$$\hat{\mathbf{g}}_t = \nabla f_t(\mathbf{w}_t)$$

where the sample $(\mathbf{x}_t, y_t)$ is uniformly at random chosen from $\mathcal{S}$. To make a fair comparison, we independently repeated each stochastic algorithm ten times and report the results averaged over ten trials.

For nonsmooth cases, we consider the $l_1$-regularized hinge loss learning problems, that is,

$$\min \lambda\|\mathbf{w}\|_1 + \frac{1}{m}\sum_{i=1}^{m} f_i(\mathbf{w}) \quad (23)$$

where $f_i(\mathbf{w}) = \max\{0, 1 - y_i\langle\mathbf{w}, \mathbf{x}_i\rangle\}$. It is well-known that the tradeoff parameter $\lambda$ in (23) influences both convergence and sparsity. In this experiment, we choose to compare our SRSG (18) with stochastic COMID [10], PA-PSG (22) and SRSG (13), in which the only difference lies in their stepsize rules. The parameters $\theta_t$ and $a_t$ of SRSG (13) are selected according to Theorem 3. The convergence of the averaged solution of COMID and individual solution of PA-PSG (22) and SRSG (18) on four different data sets (Table I) are illustrated respectively in Figs. 1–4. These convergences are all proved to attain optimal rates. From these figures, it can be observed that they have almost the same behavior.

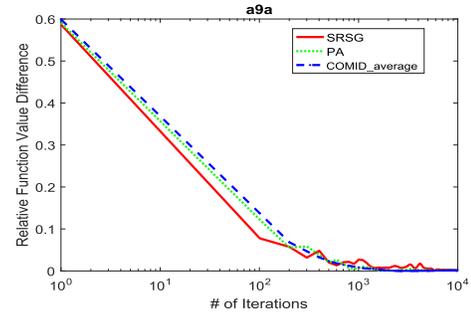

Fig. 1. Convergence on the data set A9A ($\lambda = 0.02$).

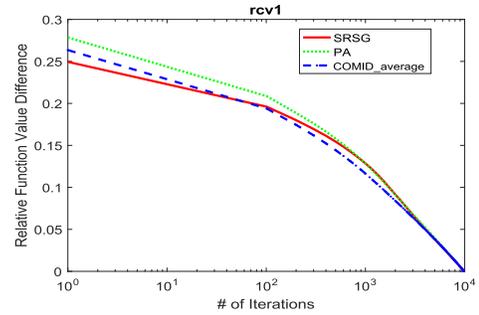

Fig. 2. Convergence on the data set RCV1 ($\lambda = 0.002$).

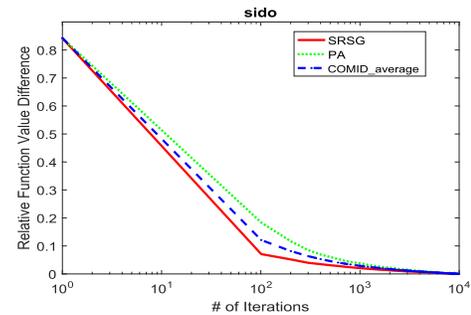

Fig. 3. Convergence on the data set SIDO ($\lambda = 0.05$).

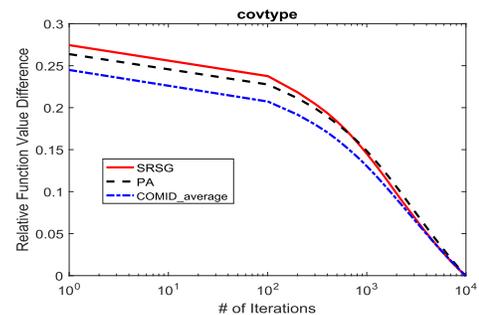

Fig. 4. Convergence on the data set COVTYPE ($\lambda = 0.01$).

The sparsity of individual output of stochastic SRSG (18) and PA-PSG (22), and the sparsity of the averaged solution of

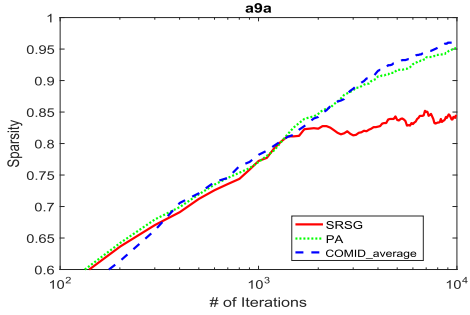

Fig. 5. Sparsity on the data set A9A ($\lambda = 0.02$).

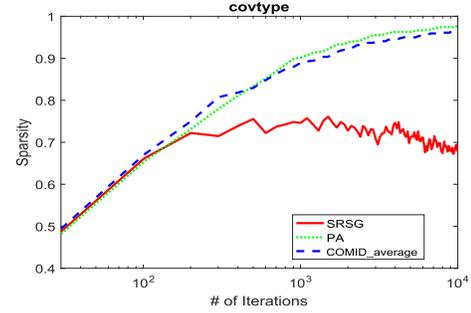

Fig. 8. Sparsity on the data set COVTYPE ($\lambda = 0.01$).

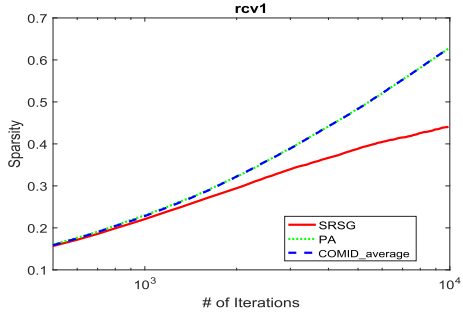

Fig. 6. Sparsity on the data set RCV1 ($\lambda = 0.002$).

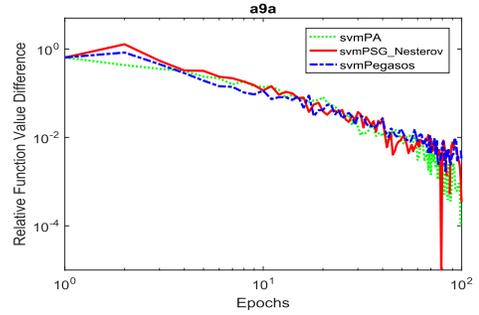

Fig. 9. Convergence on the data set A9A.

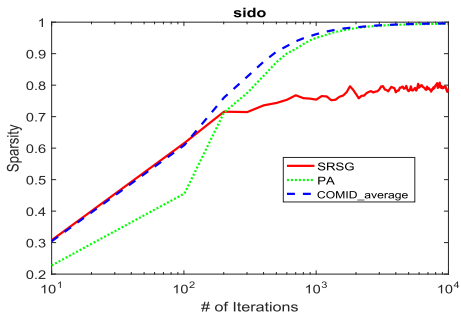

Fig. 7. Sparsity on the data set SIDO ($\lambda = 0.05$).

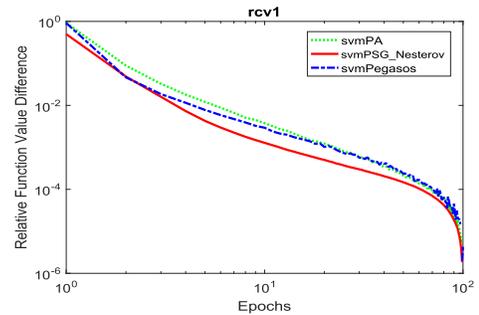

Fig. 10. Convergence on the data set RCV1.

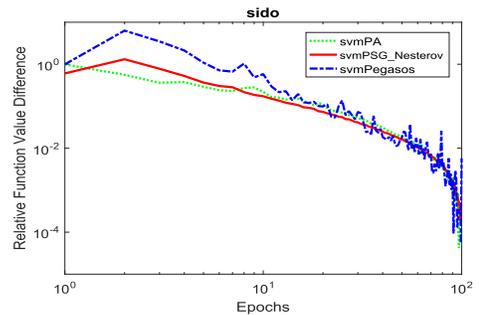

Fig. 11. Convergence on the data set SIDO.

stochastic COMID [10] on four different data sets (Table I) are illustrated respectively in Figs. 5–8. It can be seen that the individual solution of our stochastic SRSG (18) consistently has better performance while the individual output of PA-PSG (22) only derives almost the same sparsity as the averaged solution. Thus for regularized nonsmooth problems, we conclude that Nesterov's extrapolation enables PSG to have better sparsity than other strategies while attaining optimal individual convergence.

For nonsmooth but strongly convex cases, we consider standard support vector machine (SVM) problems, that is,

$$\min \frac{\lambda}{2}\|\mathbf{w}\|^2 + \frac{1}{m}\sum_{i=1}^{m} f_i(\mathbf{w}) \quad (24)$$

where $\lambda$ is a tradeoff parameter. The goal of this experiment is to verify the theoretic analysis in Section V. Note for strongly convex problems with constraints, one state-of-the-art algorithm is Pegasos and its high performance has been sufficiently illustrated in [25]. So, we compare our Nesterov-PSG (15) with Pegasos ($a_t = (1/\lambda t)$) in (8) in terms of convergence.

We also compare our Nesterov-PSG (15) with PA-PSG (12), in which their difference only lies in the stepsize rules. For all the four data sets, $\lambda = (1/n)$ ($n$ is the number of training samples) and $\mathbf{Q}$ is determined by using the strong duality theorem [25]. The parameters $\theta_t$ and $a_t$ are selected according to Theorem 1. The individual convergence of Pegasos, PA-PSG (12), and our stochastic PSG (15) on four different data sets (Table I) are illustrated, respectively, in Fig. 8–12.

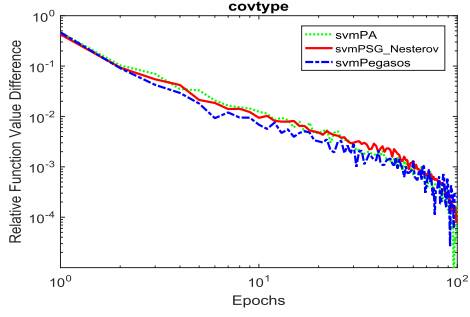

Fig. 12. Convergence on the data set COVTYPE.

From these figures, it can be observed that three algorithms have almost the same convergence. However, it should be pointed out that theoretical guarantee for optimal individual convergence of Pegasos still remains unsolved now [26]. Thus for strongly convex problems, we conclude that our Nesterov-PSG (15) has optimal individual convergence.

## VIII. CONCLUSION

In the domain of smooth optimization, a well-known fact is that Nesterov's extrapolation strategy can accelerate the convergence rate of gradientlike methods by orders of magnitude. However, as far as know, its strength in nonsmooth optimization has not been fully investigated. In this article, we prove that Nesterov's extrapolation succeeds in leading to optimal individual convergence of PSG for nonsmooth convex and strongly convex problems. These results can be regarded as making an interesting step toward the open question about SGD posed by Shamir. Besides, the final solution of our algorithm is directly generated by the gradientlike operation, which succeeds in keeping the sparsity when dealing with $l_1$-regularized learning problems.

This article only focuses on the basic optimization method PSG. Based on its equivalent reformulation, Nesterov's extrapolation can be employed in MD to achieve optimal individual convergence. Obviously, it is interesting that whether we can extend Nesterov's extrapolation to other optimization methods such as alternating direction method of multiplier (ADMM) [28] and preconditioned SGD [23], and its application like that in [32] is also expected. All these issues will be considered in our future work.

## APPENDIX

*Lemma 1:* Let $\{\mathbf{w}_t\}_{t=1}^{\infty}$ and $\{\mathbf{y}_t\}_{t=1}^{\infty}$ be generated by (13). For any $\mathbf{w} \in \mathbf{Q}$, we have

$$a_t[f(\mathbf{y}_t) - f(\mathbf{w})]$$
$$\leq \frac{1}{2}\|\mathbf{w} - \mathbf{y}_t\|^2 - \frac{1}{2}\|\mathbf{w} - \mathbf{w}_{t+1}\|^2 - \frac{1}{2}\|\mathbf{y}_t - \mathbf{w}_{t+1}\|^2$$
$$+ a_t \langle \nabla f(\mathbf{y}_t), \mathbf{y}_t - \mathbf{w}_{t+1} \rangle.$$

*Proof:* Note

$$\mathbf{w}_{t+1} = P[\mathbf{y}_t - a_t \nabla f(\mathbf{y}_t)]$$

is equivalent to

$$\mathbf{w}_{t+1} = \arg\min_{\mathbf{w} \in \mathbf{Q}} \left\{ a_t \langle \nabla f(\mathbf{y}_t), \mathbf{w} \rangle + \frac{1}{2}\|\mathbf{w} - \mathbf{y}_t\|^2 \right\}.$$

According to the convexity of $f$

$$a_t[f(\mathbf{y}_t) - f(\mathbf{w})]$$
$$\leq a_t \langle \nabla f(\mathbf{y}_t), \mathbf{y}_t - \mathbf{w} \rangle$$
$$= a_t \langle \nabla f(\mathbf{y}_t), \mathbf{y}_t \rangle - a_t \langle \nabla f(\mathbf{y}_t), \mathbf{w} \rangle - \frac{1}{2}\|\mathbf{w} - \mathbf{y}_t\|^2$$
$$+ \frac{1}{2}\|\mathbf{w} - \mathbf{y}_t\|^2. \quad (25)$$

Note the objective function $a_t \langle \nabla f(\mathbf{y}_t), \mathbf{w} \rangle + (1/2)\|\mathbf{w} - \mathbf{y}_t\|^2$ is $(1/2)$-strongly convex, then we have

$$a_t \langle \nabla f(\mathbf{y}_t), \mathbf{w} \rangle + \frac{1}{2}\|\mathbf{w} - \mathbf{y}_t\|^2$$
$$\geq a_t \langle \nabla f(\mathbf{y}_t), \mathbf{w}_{t+1} \rangle + \frac{1}{2}\|\mathbf{w}_{t+1} - \mathbf{y}_t\|^2 + \frac{1}{2}\|\mathbf{w} - \mathbf{w}_{t+1}\|^2.$$

Thus Lemma 1 is proven.

In contrast to the proof in smooth optimizations [30], $-(1/2)\|\mathbf{y}_t - \mathbf{w}_{t+1}\|^2 + a_t \langle \nabla f(\mathbf{y}_t), \mathbf{y}_t - \mathbf{w}_{t+1} \rangle$ is an extra term caused by the nonsmooth objective function. Using the technique in [10], we can incorporate these two terms into $\frac{1}{2}a_t^2 M^2$ by applying the Fenchel-Young inequality, that is,

*Lemma 2:* Let $\{\mathbf{w}_t\}_{t=1}^{\infty}$ be generated by (13). For any $\mathbf{w} \in \mathbf{Q}$, we have

$$a_t[f(\mathbf{w}_{t+1}) - f(\mathbf{w})] \leq \frac{1}{2}\|\mathbf{w} - \mathbf{y}_t\|^2 - \frac{1}{2}\|\mathbf{w} - \mathbf{w}_{t+1}\|^2 + \frac{1}{2}a_t^2 M^2$$

where $M$ is defined in Assumption 1.

*Proof:* According to Lemma 1

$$a_t[f(\mathbf{w}_{t+1}) - f(\mathbf{w})]$$
$$= a_t[f(\mathbf{w}_{t+1}) - f(\mathbf{y}_t)] + a_t[f(\mathbf{y}_t) - f(\mathbf{w})]$$
$$\leq \frac{1}{2}\|\mathbf{w} - \mathbf{y}_t\|^2 - \frac{1}{2}\|\mathbf{w} - \mathbf{w}_{t+1}\|^2 - \frac{1}{2}\|\mathbf{y}_t - \mathbf{w}_{t+1}\|^2$$
$$+ a_t \langle \nabla f(\mathbf{y}_t), \mathbf{y}_t - \mathbf{w}_{t+1} \rangle + a_t[f(\mathbf{w}_{t+1}) - f(\mathbf{y}_t)]$$

From Assumption 1, $f$ is Lipschitz with constant $M$. Then we can establish the relationship between $f(\mathbf{w}_{t+1})$ and $f(\mathbf{y}_t)$, that is,

$$f(\mathbf{w}_{t+1}) \leq f(\mathbf{y}_t) + \langle \nabla f(\mathbf{y}_t), \mathbf{w}_{t+1} - \mathbf{y}_t \rangle + M\|\mathbf{y}_t - \mathbf{w}_{t+1}\|.$$

Note

$$-\frac{1}{2}\|\mathbf{y}_t - \mathbf{w}_{t+1}\|^2 + a_t M \|\mathbf{y}_t - \mathbf{w}_{t+1}\| \leq \frac{1}{2}a_t^2 M^2.$$

Thus Lemma 2 is proven.

Note that $(1/2)\|\mathbf{w} - \mathbf{y}_t\|^2 - (1/2)\|\mathbf{w} - \mathbf{w}_{t+1}\|^2$ is not convenient for further recursion, we can use similar technique in smooth optimization [30], that is,

*Lemma 3:* Let $\{\mathbf{w}_t\}_{t=1}^{\infty}$ be generated by (13). Let

$$\mathbf{z}_t = -(\theta_t^{-1} - 1)\mathbf{w}_t + \theta_t^{-1}\mathbf{y}_t. \quad (26)$$

For any $\mathbf{w} \in \mathbf{Q}$, we have

$$a_t[f(\mathbf{w}_{t+1}) - f((1 - \theta_t)\mathbf{w}_t + \theta_t \mathbf{w})]$$
$$\leq \frac{\theta_t^2}{2}\|\mathbf{w} - \mathbf{z}_t\|^2 - \frac{\theta_t^2}{2}\|\mathbf{w} - \mathbf{z}_{t+1}\|^2 + \frac{1}{2}a_t^2 M^2. \quad (27)$$

*Proof:* Let us first replace $\mathbf{w}$ in Lemma 2 with $(1-\theta_t)\mathbf{w}_t + \theta_t\mathbf{w}$, it holds

$$a_t[f(\mathbf{w}_{t+1}) - f((1-\theta_t)\mathbf{w}_t + \theta_t\mathbf{w})]$$
$$\leq \frac{1}{2}\|(1-\theta_t)\mathbf{w}_t + \theta_t\mathbf{w} - \mathbf{y}_t\|^2$$
$$- \frac{1}{2}\|(1-\theta_t)\mathbf{w}_t + \theta_t\mathbf{w} - \mathbf{w}_{t+1}\|^2 + \frac{1}{2}a_t^2 M^2.$$

Note
$$\frac{1}{2}\left[\|(1-\theta_t)\mathbf{w}_t + \theta_t\mathbf{w} - \mathbf{y}_t\|^2 - \|(1-\theta_t)\mathbf{w}_t + \theta_t\mathbf{w} - \mathbf{w}_{t+1}\|^2\right]$$
$$= \frac{\theta_t^2}{2}\|\mathbf{w} + (\theta_t^{-1} - 1)\mathbf{w}_t - \theta_t^{-1}\mathbf{y}_t\|^2$$
$$- \frac{\theta_t^2}{2}\|\mathbf{w} + (\theta_t^{-1} - 1)\mathbf{w}_t - \theta_t^{-1}\mathbf{w}_{t+1}\|^2.$$

According to the definition of $\mathbf{z}_t$ and (13)
$$\mathbf{z}_{t+1} = -(\theta_{t+1}^{-1} - 1)\mathbf{w}_{t+1} + \theta_{t+1}^{-1}\mathbf{y}_{t+1}$$
$$= -(\theta_t^{-1} - 1)\mathbf{w}_t + \theta_t^{-1}\mathbf{w}_{t+1}. \quad (28)$$

So
$$\frac{1}{2}\left[\|(1-\theta_t)\mathbf{w}_t + \theta_t\mathbf{w} - \mathbf{y}_t\|^2 - \|(1-\theta_t)\mathbf{w}_t + \theta_t\mathbf{w} - \mathbf{w}_{t+1}\|^2\right]$$
$$= \frac{\theta_t^2}{2}\|\mathbf{w} - \mathbf{z}_t\|^2 - \frac{\theta_t^2}{2}\|\mathbf{w} - \mathbf{z}_{t+1}\|^2.$$

Therefore Lemma 3 is proven.

In smooth optimization, the constant stepsize directly enable us to recursively deal with $\|\mathbf{w} - \mathbf{z}_k\|^2 - \|\mathbf{w} - \mathbf{z}_{k+1}\|^2$. However, $a_k$ is time-varying in nonsmooth optimization. Fortunately, we can employ the technique in [35] to cope with $\sum_{k=1}^{t}(1/2a_k)[\|\mathbf{w} - \mathbf{z}_k\|^2 - \|\mathbf{w} - \mathbf{z}_{k+1}\|^2]$.

*Lemma 4:* Let $\{\mathbf{w}_t\}_{t=1}^{\infty}$ be generated by (13). For any $\mathbf{w} \in \mathbf{Q}$, we have

$$f(\mathbf{w}_{t+1}) - f(\mathbf{w})$$
$$\leq \theta_t^2 \left[ f(\mathbf{w}_1) - f(\mathbf{w}) + \sum_{k=1}^{t} \frac{1}{2a_k}\left[\|\mathbf{w}-\mathbf{z}_k\|^2 - \|\mathbf{w}-\mathbf{z}_{k+1}\|^2\right] \right.$$
$$\left. + \sum_{k=1}^{t} \frac{a_k}{2\theta_k^2}M^2 \right]. \quad (29)$$

and
$$\|\mathbf{w}^* - \mathbf{z}_{t+1}\|^2$$
$$\leq 2a_t\left[f(\mathbf{w}_1) - f(\mathbf{w}^*) + \sum_{k=1}^{t}\frac{a_k}{2\theta_k^2}M^2 + \frac{1}{2a_1}\|\mathbf{w}^* - \mathbf{z}_1\|^2\right.$$
$$\left. + \sum_{k=2}^{t}\left(\frac{1}{2a_k} - \frac{1}{2a_{k-1}}\right)\|\mathbf{w}^* - \mathbf{z}_k\|^2\right]. \quad (30)$$

*Proof:* Since $f$ is convex
$$f((1-\theta_t)\mathbf{w}_t + \theta_t\mathbf{w}) \leq (1-\theta_t)f(\mathbf{w}_t) + \theta_t f(\mathbf{w}). \quad (31)$$

Using Lemma 3
$$a_t[f(\mathbf{w}_{t+1}) - f(\mathbf{w})] \leq a_t(1-\theta_t)[f(\mathbf{w}_t) - f(\mathbf{w})] + \frac{\theta_t^2}{2}\|\mathbf{w}-\mathbf{z}_t\|^2$$
$$- \frac{\theta_t^2}{2}\|\mathbf{w}-\mathbf{z}_{t+1}\|^2 + \frac{1}{2}a_t^2 M^2.$$

Dividing both sides by $a_t\theta_t^2$ and using (13)
$$\frac{f(\mathbf{w}_{t+1}) - f(\mathbf{w})}{\theta_t^2} \leq \frac{(1-\theta_t)[f(\mathbf{w}_t) - f(\mathbf{w})]}{\theta_t^2} + \frac{a_t}{2\theta_t^2}M^2$$
$$+ \frac{1}{2a_t}\|\mathbf{w}-\mathbf{z}_t\|^2 - \frac{1}{2a_t}\|\mathbf{w}-\mathbf{z}_{t+1}\|^2$$
$$\leq \frac{f(\mathbf{w}_t) - f(\mathbf{w})}{\theta_{t-1}^2} + \frac{a_t}{2\theta_t^2}M^2$$
$$+ \frac{1}{2a_t}\|\mathbf{w}-\mathbf{z}_t\|^2 - \frac{1}{2a_t}\|\mathbf{w}-\mathbf{z}_{t+1}\|^2.$$

Recursively applying the above inequality, we can obtain
$$\frac{f(\mathbf{w}_{t+1}) - f(\mathbf{w})}{\theta_t^2} \leq \frac{f(\mathbf{w}_1) - f(\mathbf{w})}{\theta_0^2} + \sum_{k=1}^{t}\frac{a_k}{2\theta_k^2}M^2$$
$$+ \sum_{k=1}^{t}\left[\frac{1}{2a_k}\|\mathbf{w}-\mathbf{z}_k\|^2 - \frac{1}{2a_k}\|\mathbf{w}-\mathbf{z}_{k+1}\|^2\right].$$

This is
$$\frac{f(\mathbf{w}_{t+1}) - f(\mathbf{w})}{\theta_t^2}$$
$$\leq \frac{f(\mathbf{w}_1) - f(\mathbf{w})}{\theta_0^2} + \sum_{k=1}^{t}\frac{a_k}{2\theta_k^2}M^2 + \frac{1}{2a_1}\|\mathbf{w}-\mathbf{z}_1\|^2$$
$$+ \sum_{k=2}^{t}\left(\frac{1}{2a_k} - \frac{1}{2a_{k-1}}\right)\|\mathbf{w}-\mathbf{z}_k\|^2 - \frac{1}{2a_t}\|\mathbf{w}-\mathbf{z}_{t+1}\|^2.$$

Let $\mathbf{w} = \mathbf{w}^*$. Note that $\theta_0 = 1$ and $f(\mathbf{w}_{t+1}) \geq f(\mathbf{w}^*)$. Thus Lemma 4 is proven.

*Proof of Theorem 1:*
For the specific $\theta_t = (2/t+1)$ and $a_t = (1/(t+1)\sqrt{t+1})$
$$\sum_{k=1}^{t}\frac{a_k}{2\theta_k^2} = \frac{1}{8}\sum_{k=1}^{t}\sqrt{k+1} \leq \frac{3}{16}(t+1)^{\frac{3}{2}}.$$

Using (30) in Lemma 4, by simple deduction, we can prove that there exists a positive number $M_0 > 0$ such that
$$\|\mathbf{w}^* - \mathbf{z}_{t+1}\|^2 \leq M_0 \quad \forall t > 0.$$

Since the sequence $\|\mathbf{w}^* - \mathbf{z}_{t+1}\|$ is bounded, according to (29), we have
$$f(\mathbf{w}_{t+1}) - f(\mathbf{w}^*)$$
$$\leq \theta_t^2\left[f(\mathbf{w}_1) + \frac{1}{2a_t}M_0 + \sum_{k=1}^{t}\frac{a_k}{2\theta_k^2}M^2\right]$$
$$\leq O\left(\frac{1}{\sqrt{t}}\right).$$

Thus Theorem 1 is proven.

*Lemma 5:* Let $\{\mathbf{w}_t\}_{t=1}^{\infty}$ and $\{\mathbf{y}_t\}_{t=1}^{\infty}$ be generated by (15). For any $\mathbf{w} \in \mathbf{Q}$, it holds
$$a_t[f(\mathbf{y}_t) - f(\mathbf{w})]$$
$$\leq \frac{1}{2}\|\mathbf{w} - \mathbf{y}_t\|^2 + \frac{a_t}{2\theta_t}\mu\|\mathbf{w} - \mathbf{w}_t\|^2$$
$$- \frac{1}{2}\left(1 + \frac{a_t}{\theta_t}\mu\right)\|\mathbf{w} - \mathbf{w}_{t+1}\|^2 + \frac{1}{2}a_t^2 M^2.$$

*Proof:* Note that

$$\mathbf{w}_{t+1} = P\left[\frac{\theta_t}{\theta_t + a_t\mu}\mathbf{y}_t + \frac{a_t\mu}{\theta_t + a_t\mu}\mathbf{w}_t - \frac{a_t\theta_t}{\theta_t + a_t\mu}\nabla f(\mathbf{y}_t)\right]$$

is equivalent to

$$\mathbf{w}_{t+1} = \arg\min_{\mathbf{w}\in\mathbf{Q}}\left\{a_t\langle\nabla f(\mathbf{y}_t),\mathbf{w}\rangle + \frac{1}{2}\|\mathbf{w} - \mathbf{y}_t\|^2 + \frac{a_t}{2\theta_t}\mu\|\mathbf{w} - \mathbf{w}_t\|^2\right\}.$$

In strongly convex cases, the parameter of strong convexity of the objective function in each subproblem is $1 + (a_t/\theta_t)\mu$. Following the proof of Lemmas 1 and 2, Lemma 5 is proven.

*Lemma 7:* Let $\{\mathbf{w}_t\}_{t=1}^{\infty}$ be generated by (15). Let $\mathbf{z}_t = -(\theta_t^{-1} - 1)\mathbf{w}_t + \theta_t^{-1}\mathbf{y}_t$, it holds

$$f(\mathbf{w}_{t+1}) - f(\mathbf{w}^*)$$
$$\leq \theta_t^2\left[f(\mathbf{w}_1) - f(\mathbf{w}^*) + \sum_{k=1}^{t}\frac{a_k}{2\theta_k^2}M^2 + \frac{1}{2}\sum_{k=1}^{t}\left[\frac{1}{a_k}\|\mathbf{w}^* - \mathbf{z}_k\|^2 - \left(\frac{1}{a_k} + \frac{\mu}{\theta_k}\right)\|\mathbf{w}^* - \mathbf{z}_{k+1}\|^2\right]\right].$$

*Proof:* Like the proof of Lemma 3, we first replace $\mathbf{w}$ in Lemma 6 with $(1-\theta_t)\mathbf{w}_t + \theta_t\mathbf{w}$. Unlike Lemma 3, the term $(a_t/2\theta_t)\mu\|\mathbf{w} - \mathbf{w}_t\|^2$ caused by the strong convexity in Lemma 6 becomes

$$\frac{a_t\theta_t}{2}\mu\|\mathbf{w} - \mathbf{w}_t\|^2.$$

Since $f$ is strongly convex, we use (14) to replace (31) in the proof of Lemma 4 and get

$$a_t[f(\mathbf{w}_{t+1}) - f(\mathbf{w})]$$
$$\leq a_t(1-\theta_t)[f(\mathbf{w}_t) - f(\mathbf{w})] + \frac{\theta_t^2}{2}\|\mathbf{w} - \mathbf{z}_t\|^2$$
$$+ \frac{a_t\theta_t}{2}\mu\|\mathbf{w} - \mathbf{w}_t\|^2 - \frac{\theta_t^2}{2}\left(1 + \frac{a_t}{\theta_t}\mu\right)\|\mathbf{w} - \mathbf{z}_{t+1}\|^2$$
$$- \frac{a_t\theta_t(1-\theta_t)}{2}\mu\|\mathbf{w} - \mathbf{w}_t\|^2 + \frac{1}{2}a_t^2 M^2.$$

This is

$$\frac{f(\mathbf{w}_{t+1}) - f(\mathbf{w})}{\theta_t^2}$$
$$\leq \frac{(1-\theta_t)[f(\mathbf{w}_t) - f(\mathbf{w})]}{\theta_t^2} + \frac{1}{2}\mu\|\mathbf{w} - \mathbf{w}_t\|^2$$
$$+ \frac{1}{2}\left[\frac{1}{a_t}\|\mathbf{w} - \mathbf{z}_t\|^2 - \left(\frac{1}{a_t} + \frac{\mu}{\theta_t}\right)\|\mathbf{w} - \mathbf{z}_{t+1}\|^2\right] + \frac{a_k}{2\theta_k^2}M^2.$$

When $\mathbf{w} = \mathbf{w}^*$, the strong convexity of $f$ implies

$$f(\mathbf{w}_t) - f(\mathbf{w}^*) \geq \frac{1}{2}\mu\|\mathbf{w}^* - \mathbf{w}_t\|^2.$$

Now, we have

$$\frac{f(\mathbf{w}_{t+1}) - f(\mathbf{w}^*)}{\theta_t^2}$$
$$\leq \frac{(1-\theta_t+\theta_t^2)[f(\mathbf{w}_t) - f(\mathbf{w}^*)]}{\theta_t^2} + \frac{a_k}{2\theta_k^2}M^2$$
$$+ \frac{1}{2}\left[\frac{1}{a_t}\|\mathbf{w}^* - \mathbf{z}_t\|^2 - \left(\frac{1}{a_t} + \frac{\mu}{\theta_t}\right)\|\mathbf{w}^* - \mathbf{z}_{t+1}\|^2\right]$$
$$\leq \frac{f(\mathbf{w}_t) - f(\mathbf{w}^*)}{\theta_{t-1}^2} + \frac{a_k}{2\theta_k^2}M^2$$
$$+ \frac{1}{2}\left[\frac{1}{a_t}\|\mathbf{w}^* - \mathbf{z}_t\|^2 - \left(\frac{1}{a_t} + \frac{\mu}{\theta_t}\right)\|\mathbf{w}^* - \mathbf{z}_{t+1}\|^2\right].$$

Thus Lemma 7 is proven.

*Proof of Theorem 2:* To recursively deal with the term $(1/a_k)\|\mathbf{w}^* - \mathbf{z}_k\|^2 - ((1/a_k) + (\mu/\theta_k))\|\mathbf{w}^* - \mathbf{z}_{k+1}\|^2$, we use specific $a_t$ and $\theta_t$.

It is easy to find that $\theta_t$ in Theorem 1 satisfies $\theta_t \in (0, 1]$ and $(1 - \theta_t + \theta_t^2/\theta_t^2) \leq (1/\theta_{t-1}^2)$

$$\frac{1}{2}\sum_{k=1}^{t}\left[\frac{1}{a_k}\|\mathbf{w}^* - \mathbf{z}_k\|^2 - \left(\frac{1}{a_k} + \frac{\mu}{\theta_k}\right)\|\mathbf{w}^* - \mathbf{z}_{k+1}\|^2\right]$$
$$= \frac{1}{2}\sum_{k=1}^{t}\left[\frac{\mu k^2}{3}\|\mathbf{w}^* - \mathbf{z}_k\|^2 - \mu\left(\frac{k^2}{3} + \frac{k+1}{3}\right)\|\mathbf{w}^* - \mathbf{z}_{k+1}\|^2\right]$$
$$\leq \frac{1}{2}\sum_{k=1}^{t}\left[\frac{\mu k^2}{3}\|\mathbf{w}^* - \mathbf{z}_k\|^2 - \frac{\mu}{3}(k^2 + 2k + 1)\|\mathbf{w}^* - \mathbf{z}_{k+1}\|^2\right]$$
$$\leq \frac{1}{2}\sum_{k=1}^{t}\left[\frac{\mu k^2}{3}\|\mathbf{w}^* - \mathbf{z}_k\|^2 - \frac{\mu}{3}(k+1)^2\|\mathbf{w}^* - \mathbf{z}_{k+1}\|^2\right]$$
$$\leq \frac{\mu}{6}\|\mathbf{w}^* - \mathbf{z}_1\|^2.$$

Note

$$\sum_{k=1}^{t}\frac{a_k}{2\theta_k^2} \leq \sum_{k=1}^{t}\frac{1}{6\mu}\left(1 + \frac{1}{k}\right)^2 \leq \frac{2t}{3\mu}.$$

Thus Theorem 2 is proven.

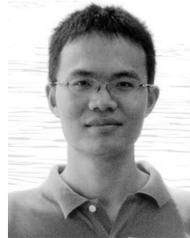

**Wei Tao** is currently pursuing the Ph.D. degree with the Army Engineering University of PLA, Nanjing, China.

His current research interests include machine learning, optimization algorithms, and network security.

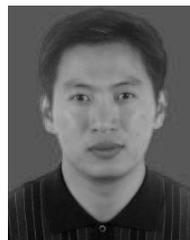

**Zhisong Pan** received the Diploma degree in computer science and technology from PLA Information Engineering University, Zhengzhou, China, in 1996, and the Ph.D. degree in computer science and technology from the Nanjing University of Aeronautics and Astronautics, Nanjing, China, in 2003.

He is currently a Professor with the Army Engineering University of PLA, Nanjing, China. His current research interests include deep learning, machine learning, and pattern recognition.

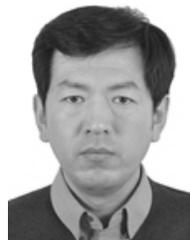

**Gaowei Wu** received the Ph.D. degree from the Institute of Automation, Chinese Academy of Sciences, Beijing, China, in 2003.

From June 2003 to June 2006, he was a Post-Doctoral Fellow with the Institute of Computing Technology, Chinese Academy of Sciences, where he is currently an Associate Professor with the Institute of Automation. His current research interests include big data and statistical learning theory.

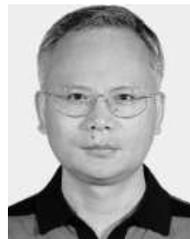

**Qing Tao** received the M.S. degree in mathematics from Southwest University, Chongqing, China, in 1989, and the Ph.D. degree from the University of Science and Technology of China, Hefei, China, in 1999.

From June 1999 to June 2001, he was a Post-Doctoral Fellow with the University of Science and Technology of China, Hefei, China. From June 2001 to 2003, he was a Post-Doctoral Fellow with the Institute of Automation, Chinese Academy of Sciences, Beijing, China, where he was a Professor with the Institute of Automation, from 2004 to 2012. He is currently a Professor with the Army Academy of Artillery and Air Defense, Hefei. His current research interests include applied mathematics, machine learning, and optimization theory.